\documentclass[letterpaper]{article}
\usepackage{arxiv}

\usepackage[utf8]{inputenc} % allow utf-8 input
\usepackage[T1]{fontenc}    % use 8-bit T1 fonts
\usepackage{hyperref}       % hyperlinks
\usepackage{url}            % simple URL typesetting
\usepackage{booktabs}       % professional-quality tables
\usepackage{amsfonts}       % blackboard math symbols
\usepackage{nicefrac}       % compact symbols for 1/2, etc.
\usepackage{microtype}      % microtypography
\usepackage{lipsum}
\usepackage{graphicx}
\usepackage{multicol}
\graphicspath{ {./images/} }
\usepackage{amsmath}
\usepackage[ruled,vlined]{algorithm2e}
\usepackage{xcolor}
\usepackage{caption}

\title{DiffusionNet: Accelerating the solution of  Time-Dependent partial differential equations  using deep learning.}

\author{

 Mahmoud Asem \\
  Department of Mechanical engineering ,KAIST \\
  \texttt{asem00@kaist.ac.kr} \\

}

\begin{document}

\maketitle
\begin{multicols}{2}
\begin{abstract}

We present our deep learning framework to solve and accelerate the Time-Dependent partial differential equation's solution of one and two spatial dimensions. We demonstrate DiffusionNet solver by solving the 2D transient heat conduction problem with Dirichlet boundary conditions. The model is trained on solution data calculated using the Alternating direction implicit method. We show the model's ability to predict the solution from any combination of seven variables: the starting time step of the solution, initial condition, four boundary conditions, and a combined variable of the time step size, diffusivity constant, and grid step size. To improve speed, we exploit our model capability to predict the solution of the Time-dependent PDE after multiple time steps at once  to improve the speed of solution by dividing the solution into parallelizable chunks. 
We try to build a flexible architecture capable of solving a wide range of partial differential equations with minimal changes. We demonstrate our model flexibility by applying our model with the same network architecture used to solve the transient heat conduction to solve the Inviscid Burgers equation and Steady-state heat conduction, then compare our model performance against related studies. We show that our model reduces the error of the solution for the investigated problems.

\end{abstract}

% keywords can be removed

\section{Introduction}

\subsection{Motivation}
The importance of partial differential equations stems from the fact that fundamental physical laws are formulated in partial differential equations; examples include the Schrödinger equation, Heat equation, Navier-Stokes equations, and linear elasticity equation. 
Partial differential equations are solved analytically and numerically. Numerical methods discretize the solution domain and construct algebraic equations that are solved either analytically or iteratively. Numerical methods include the finite difference method, finite volume method, finite element method, and meshless methods. The solution of the equations becomes costly if the number of equations is high. Moreover, PDE solutions can be distinctively different, and no general approach is applicable to all kinds of PDE.

These reasons, among others, are a prime motivation for utilizing machine learning techniques.
The research for solving partial differential equations with machine learning can be divided into data-driven and data-free; in data-driven methods, the data is generated experimentally or numerically and then fed to the model to learn the underlying relations \cite{ranade2020discretizationnet}.In data-free methods, the neural network yields solution by utilizing loss formulation to constrain the partial governing differential. In this paper, we focus on the data-driven methods of solutions to solve three partial differential equations: (1) Transient heat conduction, (2) Inviscid Burgers’ equation, (3) Steady-state heat conduction.

\subsection{Related work}

Many deep learning-based approaches have been developed to solve numerous heat and fluid transfer problems. In the following, we briefly introduce the efforts to solve the related partial differential equations using deep learning. For the heat conduction problem,\cite{sharma2018weakly}Used U-Net architecture to infer the Laplace equation's solution with Dirichlet boundary condition to estimate the temperature distribution over a flat plate. \cite{farimani2017deep} Used conditional generative adversarial networks (cGAN) to generate the solution of steady-state heat conduction problem and steady-state fluid flow in a two-dimensional domain problem via a data-driven paradigm. The authors generated Their dataset using the finite difference method for different domain sizes, geometries, and temperature boundary conditions.\cite{zakeri2019deep} Used the Finite volume method to discretize the Laplace equation and generated 100000 different cases of  19 elements of geometry and temperature values, then fed the data to a 21-neuron network to solve the temperature distribution problem.\cite{edalatifar2020using} Introduced Mean of the maximum of square error MMaSE as loss function for the steady-state heat transfer problem, then used data sets of different geometries to train the convolutional autoencoder network to infer the Laplace equation's solution without an iterative computational step.

\subsection{Our contribution}
We try to present a flexible  architecture capable of solving a wide range of partial differential equations by applying our model to solve time-dependent partial differential equations in one and two spatial domains and a time-independent partial differential equation in two spatial dimensions. We discuss the different problems the model capable of solving using the proposed architecture and training scheme. We propose to use the Convolutional-Convolutional LSTM autoencoder network to capture the spatial and temporal features respectively and to extend our model capability to solve the spatiotemporal problems. 
We propose deep learning and physics solver coupled framework to speed up the computation by exploiting the architecture and training scheme to enable our model to solve in parallell. We apply this method for two different time-dependent partial differential equations. 
Moreover, we compare our solution method for a one-dimensional time-dependent partial differential equation with baseline algorithms and show that our model can be used to reduce the error . 
The paper is organized as follows; we present the necessary background , Then we discuss the details of the network architecture, training scheme, and solution scheme in the third section. In the fourth section, we present the experimental results and discuss the transient heat conduction solution. In the fifth and sixth sections, We use data provided by previous studies to train our model to solve for (1) Inviscid Burgers’ equation, (2) Steady-state heat conduction, and compare our relative performance, respectively. Finally, we conclude our study with improvements for further work

\section{Background}

\subsection{Heat transport}
\label{sec:headings}
Diffusion processes are essential because they are exhibited in many applications that span multiple disciplines across science and engineering. The diffusion process is exhibited in a system that is not in equilibrium; examples include heat diffusion from hot to cold, diffusion of perfume in a closed room, or momentum diffusion in a fluid by viscous effects. Diffusion is  accompanied by an increase in the entropy of the universe.Entropy is an integral part of the diffusion process, For the heat diffusion from hot to cold, the spontaneous process is exhibited because The hot surroundings increase the cold system molecular micro-states such that the sum of system entropy and surrounding entropy is positive \cite{lienhard2005heat}. We focus our analysis of the PDE in the two spatial domain on the transient heat conduction in a solid problem as a representative of the diffusion processes.

An equation is referred to as a PDE if it involves partial derivatives of an unknown function of two or more independent variables \cite{chapra2010numerical}. The order of PDE  is the highest order partial derivative in the equation. The  PDE is considered linear If it is linear in the unknown function and its derivatives with coefficients depending only on the independent variables. In this study, we focus on parabolic second-order linear equations represented by 
\begin{equation} \label{30.1}
    \frac{\partial T}{\partial t } =k\frac{\partial^2{T}}{\partial {x}^2}
\end{equation}

\subsection{Solving the transient Heat conduction equation}
In this section, we introduce the different solution approaches for the heat conduction problem solution in one and two spatial dimensions.

\subsubsection{Explicit method }
This method approximates the equation second derivative in space by centered finite-divided difference and first derivative by forward finite-divided difference.
\begin{equation} \label{30.2}
    \frac{\partial^2{T}}{\partial{t}^2} = \frac{T_{i+1}^{l} -2T_{i}^{l} + T_{i-1}^{l}}{(\Delta x)^2}
\end{equation}

\begin{equation} \label{30.3}
    \frac{\partial{T}}{\partial{t}} = \frac{T_{i}^{l+1}-T_{i}^{l}}{\Delta t} 
\end{equation}

Substituting Eq. (\ref{30.2}) and Eq. (\ref{30.3}) into Eq. (\ref{30.1}) results in 
\begin{equation} \label{30.4}
  k  \frac{T_{i+1}^{l} -2T_{i}^{l} + T_{i-1}^{l}}{(\Delta x)^2} = \frac{T_{i}^{l+1}-T_{i}^{l}}{\Delta t}
\end{equation}

We can then solve for $T_{i^{l+1}}$ , 

\begin{equation}\label{30.5}
    T_{i}^{l+1} = T_{i}^{l} + \lambda (T_{i+1}^{l} - 2T_{i}^{l}+T_{i-1}^{l} ) \ \ \  
\end{equation}
    \[\lambda = k \frac{\Delta t}{(\Delta x)^2} \]
    
    solution is convergent and stable if $\lambda \leq 1/2$ \cite{chapra2010numerical} , \cite{carnahanj}  
    
\subsubsection{Implicit method}

Implicit methods are employed for better stability than explicit methods. Implicit methods differ from the explicit method such that the spatial derivative is approximated at the next time step. 

\begin{equation} \label{30.7}
    \frac{\partial^2{T}}{\partial{t}^2} = \frac{T_{i+1}^{l+1} - 2T_{i}^{l+1} + T_{i-1}^{l+1}}{(\Delta x)^2}
\end{equation}

Substituting the previous approximation  in Eq (\ref{30.1})  with Eq (\ref{30.3})  yields Equations for interior nodes, first interior node, and last interior node $(i=m)$ respectively.

\begin{align}\label{implicit}
&- \lambda T^{l+1}_{i-1} + (1+ 2\lambda)T_{i}^{l+1} - \lambda T_{i+1}^{l+1} = T_{i}^{l} \\
&(1+2 \lambda) T_{1}^{l+1} - \lambda T_{2} = T_{1}^l + \lambda f_0(t^{l+1}) \\
& - \lambda T_{m-1}^{l+1} + (1+ 2 \lambda) T_{m}^{l+1} = T_{m}^{l} + \lambda f_{m+1}(t^{l+1}) 
\end{align}

where $f_0(t^{l+1})$ expresses the boundary condition as a function of time.The equations above are written as a set of linear algebraic equations in m unknowns, where m denotes the number of interior nodes of the solution grid.

\subsubsection{Crank-Nicolson method}

Crank-Nicolson method is an alternative  to the implicit method that is second-order accurate spatially and temporally. In this method, the spatial second derivative is averaged between the time steps

\begin{gather}\label{crank}
\frac{\partial T}{\partial t} = \frac{T_{i}^{l+1} - T_{i}^{l}}{\Delta t} \\
\frac{\partial^2{T}}{\partial{x}^2} = \frac{1}{2} [ \frac{T_{i+1}^{l} - 2T_{i}^{l}  + T_{i+1}^{l}}{(\Delta x)^2} + \frac{T_{i+1}^{l+1} - 2T_{i}^{l+1} + T_{i-1}^{l+1}}{(\Delta x)^2}] \notag \\
\end{gather}

Substituting Eq.(\ref{crank})  to Eq. (\ref{30.1}) yields the following equations for the first interior node , middle interior nodes and last interior node $(i=m)$ respectively,

\begin{equation}
 \begin{split}
    & 2(1+\lambda)  T_{1}^{l+1}  - \lambda T_{2}^{l+1} = \\ 
    & \lambda f_{0}(t^{l}) + 2(1-\lambda)T_{1}^{l}  + \lambda T_{2}^{l} + \lambda f_{0}(t^{l+1}) 
\end{split}
\end{equation}

\begin{equation}
  \begin{split}
  - &  \lambda T_{i-1}^{l+1} +  2 (1  +\lambda) T_{i}^{l+1} - \lambda T_{i+1}^{l+1} = \\ 
    & \lambda T_{i-1}^{l} +  2 (1  -\lambda)T_{i}^{l} + \lambda T_{i+1}^{l} 
\end{split}  
\end{equation}

\begin{equation}
  \begin{split} 
 -   \lambda T_{m-1}^{l+1}  &+ 2 (1+\lambda)T_{m}^{l+1}  =  \\
     \lambda f_{m+1}(t^{l}) +  2(1- \lambda)T_{m}^{l} & + \lambda T_{m-1}^{l} + \lambda f_{m+1}(t^{l+1}) 
\end{split}  
\end{equation}

\subsubsection{Alternating direction implicit method}

For the parabolic equation in two spatial dimensions, 

\begin{equation}
    \frac{\partial T}{\partial t} = k (\frac{\partial^2T}{\partial x^2} + \frac{\partial^2T}{\partial y^2}) 
\end{equation}

using the Crank Nicolson  will result in $m \times n$  simultaneous equations. Instead, the Alternating direction implicit  solves the equations efficiently using tridiagonal matrices. The solution time step is progressing at half step, for the first step, the spatial second derivative is approximated as

\begin{align}
       & \frac{T_{i,j}^{l+1/2}-T_{i,j}^{l}}{\Delta t/2}   = \notag \\
       & k [\frac{T_{i+1,j}^{l} - 2T_{i,j}^{l}   + T_{i-1,j}^{l}}{(\Delta x)^2} + \frac{T_{i,j+1}^{l+1/2} - 2T_{i,j}^{l+1/2}+T_{i,j-1}^{l+1/2}}{(\Delta y )^2 }] 
\end{align}

Simplifying and rearranging for a square gird ($\Delta x = \Delta y$) ,

\begin{align}
- & \lambda T_{i,j-1}^{l+1/2} + 2(1  + \lambda) T_{i,j}^{l+1/2} - \lambda T_{i,j+1}^{l+1/2} =  \notag \\
  & \lambda T_{i-1,j} + 2 (1  - \lambda)T_{i,j}^{l} + \lambda T_{i+1,j}^{l} 
\end{align}

For the second half time step ,

\begin{align}
- & \frac{T_{i,j}^{l}-T_{i,j}^{l+1/2}}{\Delta t/2}  =  \notag \\
  & k [\frac{T_{i+1,j}^{l+1} - 2T_{i,j}^{l+1} + T_{i-1,j}^{l+1}}{(\Delta x)^2}   +   \frac{T_{i,j+1}^{l+1/2} -2T_{i,j}^{l+1/2}+T_{i,j-1}^{l+1/2}}{(\Delta y )^2 }] 
\end{align}

Simplifying and rearranging yields ,

\begin{align}
- & \lambda T_{i-1,j}^{l+1}+2 (1+ \lambda) T_{i,j}^{l+1} - \lambda T_{i+1,j}^{l+1} =  \notag \\
  & \lambda T_{i,j-1}^{l+1/2} + 2(1-\lambda)T_{i,j}^{l+1/2} + \lambda T_{i,j+1}^{l+1/2} 
\end{align}

The spatial derivative approximation alternates between explicit and implicit form for the first half step and second half step, respectively. The form of the previous equations is of the tridiagonal system; thus it can be solved efficiently.

\subsection{Convolutional LSTM Background}

\subsubsection{Long short term memory}
Long short term memory (LSTM) is a recurrent neural network variant introduced to overcome the limitation of Recurrent neural network ability to utilize information in long sequences due to vanishing gradients.
LSTM introduces a recurrent forget gate to prevent the backpropagated errors from exploding or vanishing. We introduce the formulation of the LSTM structure briefly \cite{hochreiter1997long} , \cite{hochreiter1996lstm}.

\begin{align}\label{lstm}
&i_t = \sigma(W_{xi}x_t+W_{hi}h_{t-1}+W_{ci}c_{t-1}+b_i) \notag \\
&f_t = \sigma(W_{xf}x_t+W_{hf}h_{t-1}+W_{cf}c_{t-1}+b_f) \notag \\
&\tilde{c_t} = \tanh(W_{xc}x_t+W_{hc}h_{t-1}+b_c) \notag \\
&c_t= f_t \circ c_{t-1} + i_t \circ \tilde{c_t} \notag \\
&o_t = \sigma(W_{xo}x_t+W_{ho}h_{t-1}+W_{co} \circ c_t+b_o) \notag \\
&h_t = o_t \circ \tanh(c_t),
\end{align}

Equation  (\ref{lstm}.1) is the forget gate layer. Forget gate layer decides the information that discarded from the cell state using the sigmoid layer. (\ref{lstm}.2) and (\ref{lstm}.3) Equations to decide what information to store in the cell state, (\ref{lstm}.4) is the input gate layer which decides the values to be updated, (\ref{lstm}.3) tanh layer creates a vector of values candidates for addition. (\ref{lstm}.4) combine old and new states.(\ref{lstm}.5) and (\ref{lstm}.6) decides the output at the next time step. The variable $x_t$ , $h_t$ $C_t$ $W$ denotes the input vector, the hidden state, denotes the cell state at time $t$ and the trainable weight matrices. $\circ$ denotes the Hadamard product.$b$ is the bias vectors

\subsubsection{Convolutional LSTM }
\cite{xingjian2015convolutional} proposed an LSTM variant by introducing the convolutional structures in the input-to-state and state-to-state transition in the LSTM . where $*$ represents convolution operator.

\begin{align}\label{convlstm}
&f_t = \sigma(W_{xf}*X_t+W_{hf}*H_{t-1}+W_{cf} \circ C_{t-1}+b_f) \notag \\
&i_t = \sigma(W_{xi}*X_t+W_{hi}*H_{t-1}+W_{ci} \circ C_{t-1}+b_i) \notag \\
&\tilde{C_t} =\tanh(W_{xc}*X_t+W_{hc}*H_{t-1}+b_c)  \notag\\
&C_t = f_t \circ C_{t-1} + i_t \circ \tilde{C_t} \notag \\
&o_t = \sigma(W_{xo}*X_t+W_{ho}*H_{t-1}+W_{co} \circ C_t+b_o) \notag \\
&H_t = o_t \circ \tanh(C_t),
\end{align}

\section{DiffusionNet}

\subsection{Data representation}

\begin{center}
\includegraphics[width=3in]{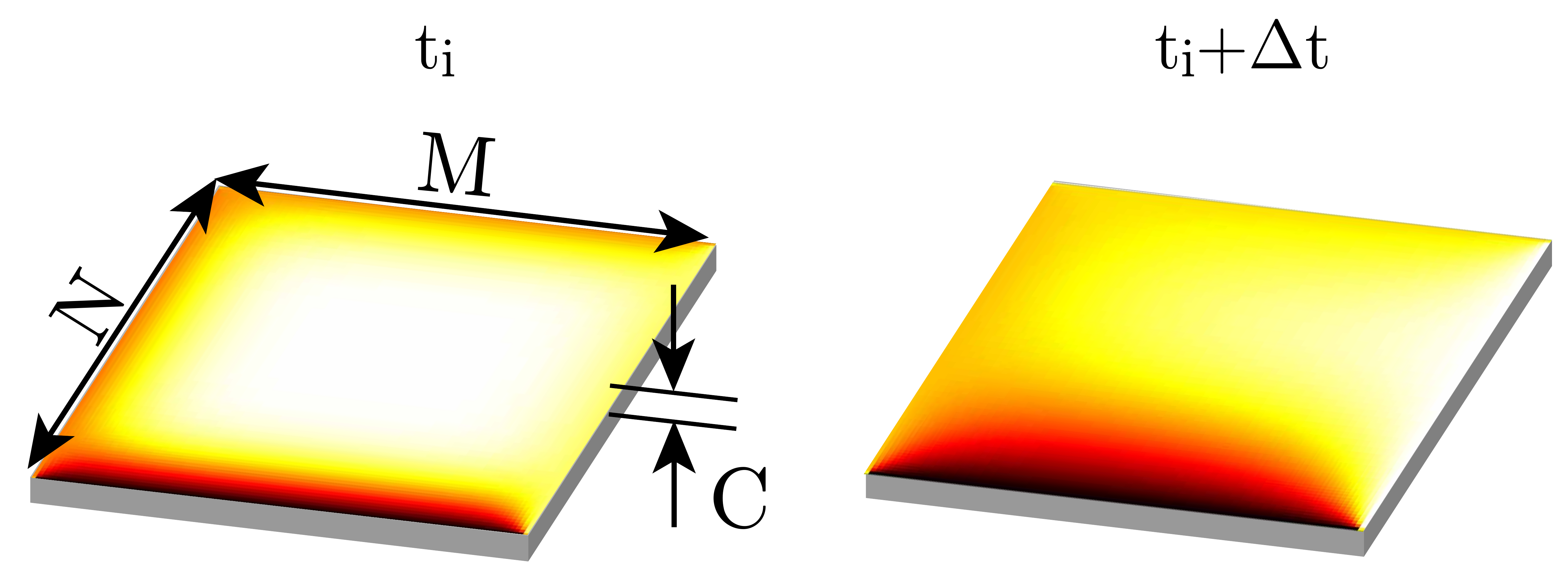}
\captionsetup{justification=centering}
\captionof{figure}{Data representation in DiffusionNet input and output.ti,N,M and C represent time step,rows , columns and channels dimensions respectively.\emph{Left} Input 4D tensor of shape $1 \times N \times M \times 1$. at time $t_i$. \emph{Right} output 4D tensor of shape $1 \times N \times M \times C$ at time $t_i + P$}    
\end{center}

The model takes a four-dimensional tensor as an input of the shape of $time steps(T) \times rows(N) \times columns(M) \times channels(C)$.The input can be considered as a video of a single frames, where each pixel at location $(i,j)$ corresponds to the value of the variable at the same location in the solution mesh. For our problem, Our quantity of interest is Temperature. Thus the single variable maps to a single channel. To track more than a single quantity of interest ( ex, Pressure, and velocity distribution in the Navier-Stokes equation ), we concatenate the grid solution along the channel dimension axis. We preprocess the input and target data by using standardization. We standardize the input and output data by subtracting the mean and dividing by the standard deviation of the entire data points.Standardization re scale the data distribution to have zero mean and standard deviation = 1.

\subsection{Network design and architecture}

\begin{center}
\includegraphics[width=\columnwidth]{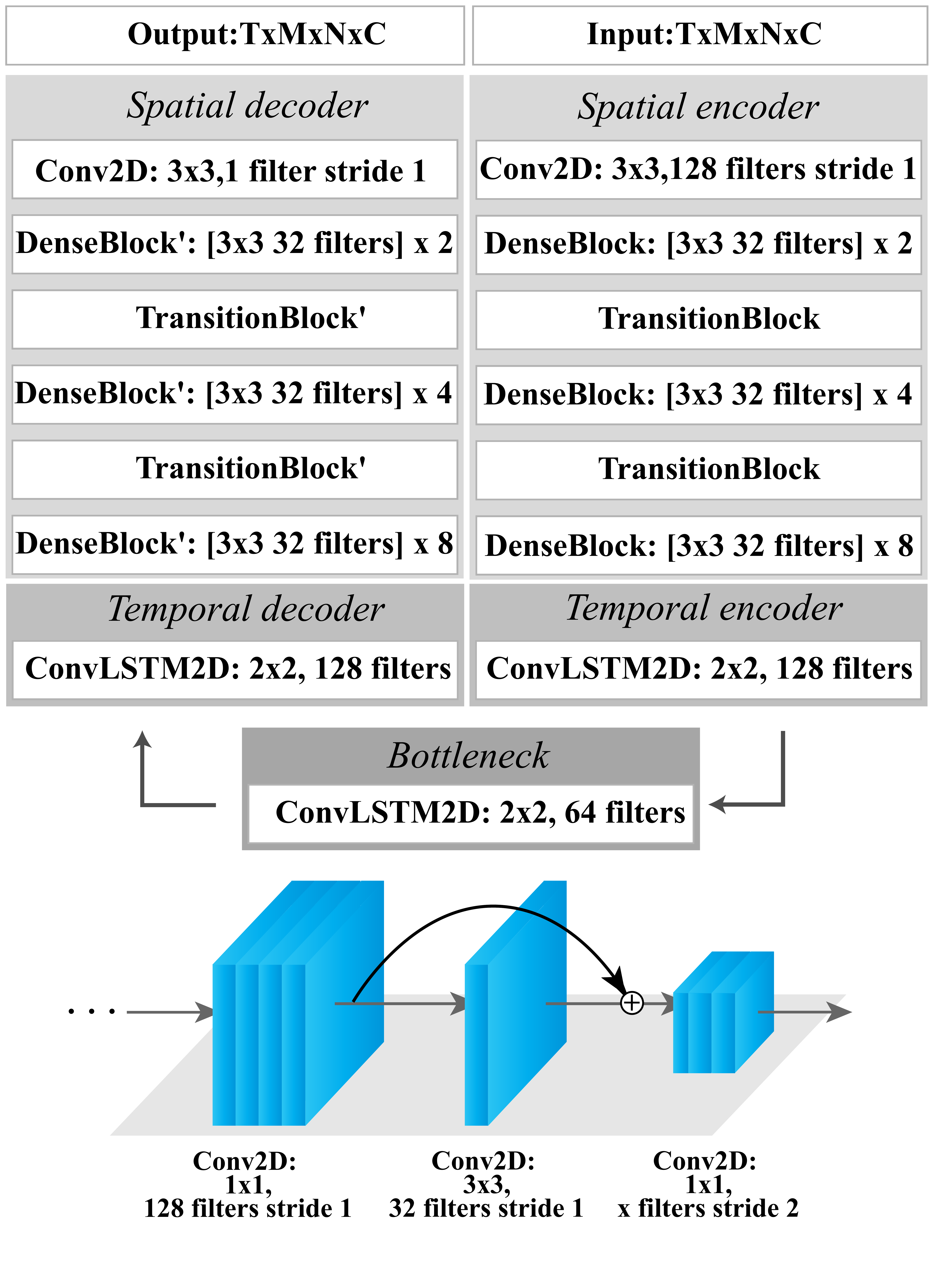}
\captionsetup{justification=centering}
\captionof{figure}{DiffusionNet network architecture.(\emph{Top}) full network architecture.(\emph{Bottom}) a DenseBlock followed by a Transition block}      
\end{center}

DiffusionNet network is a spatiotemporal autoencoder composed of locally connected convolution, deconvolution, and Convolutional LSTM layers. This type of layer connection enables arbitrary spatial dimensions and temporal dimensions as input. We use the Autoencoder structure for our network that is commonly used in Dimensionality reduction, image compression, Feature extraction, Recommendation systems, Sequence to sequence prediction, image denoising, and image generation. The structure of the autoencoder is composed of an encoder layer, bottleneck, and decoder layers. The encoder encodes and reduces the input data's dimensionality, while The bottleneck layer contains the lowest dimensional representation of the input data. The decoder reconstructs the output data from the encoded data. An autoencoder output has the same dimensions as the input.

For spatial feature extraction, we adopt Densely Connected Convolutional Networks \cite{xingjian2015convolutional}  with few modifications.DenseBlocks are utilized for two reasons: (1) It Possesses fewer parameters than a traditional convolutional network, (2) Offers improved gradient flow throughout the networks.DenseNet achieves the previous merits by connecting all the convolutional layers directly to each other, such that each layer receives the feature maps from all the subsequent layers \cite{xingjian2015convolutional}.

We adopt the direct connection and the inheritance of feature maps to implement a modified version of the dense block. Our modified DenseBlock is comprised of a bottleneck $1 \times 1 $ convolution layer of $4f$ filters  followed by $3 \times 3 $ convolutional layer of $f$ filters activated by LeakyReLU.The transition block reduces the spatial dimension of the input after each dense block by half. Our implementation of the Transition block comprises a single convolutional $1 \times 1$ layer with a stride of 2 and $x$ filters, where $x$ is defined to be half the number of the subsequent feature maps to improve the model compactness further. We replace the convolutional layer with a deconvolutional layer for the decoder and denote the deconvolution block variants by an apostrophe . For temporal feature extraction, we utilize the Convolutional LSTM layer; we choose Convolutional LSTM because it outperforms traditional FC-LSTM in its capability to learn the spatiotemporal features \cite{DBLP:journals/corr/HuangLW16a}
Throughout this paper , We initialized the layers with glorot uniform with Adam optimizer of initial learning rate = 0.001 and employed a learning rate scheduler that reduces the learning rate to half when the $\Delta MAE \ loss < 0.0001$ for three consecutive epochs. We found that combining the previous techniques with the MAE loss function is optimal for our problem.

\subsection{Solution acceleration scheme}
\begin{center}
\includegraphics[width=\columnwidth]{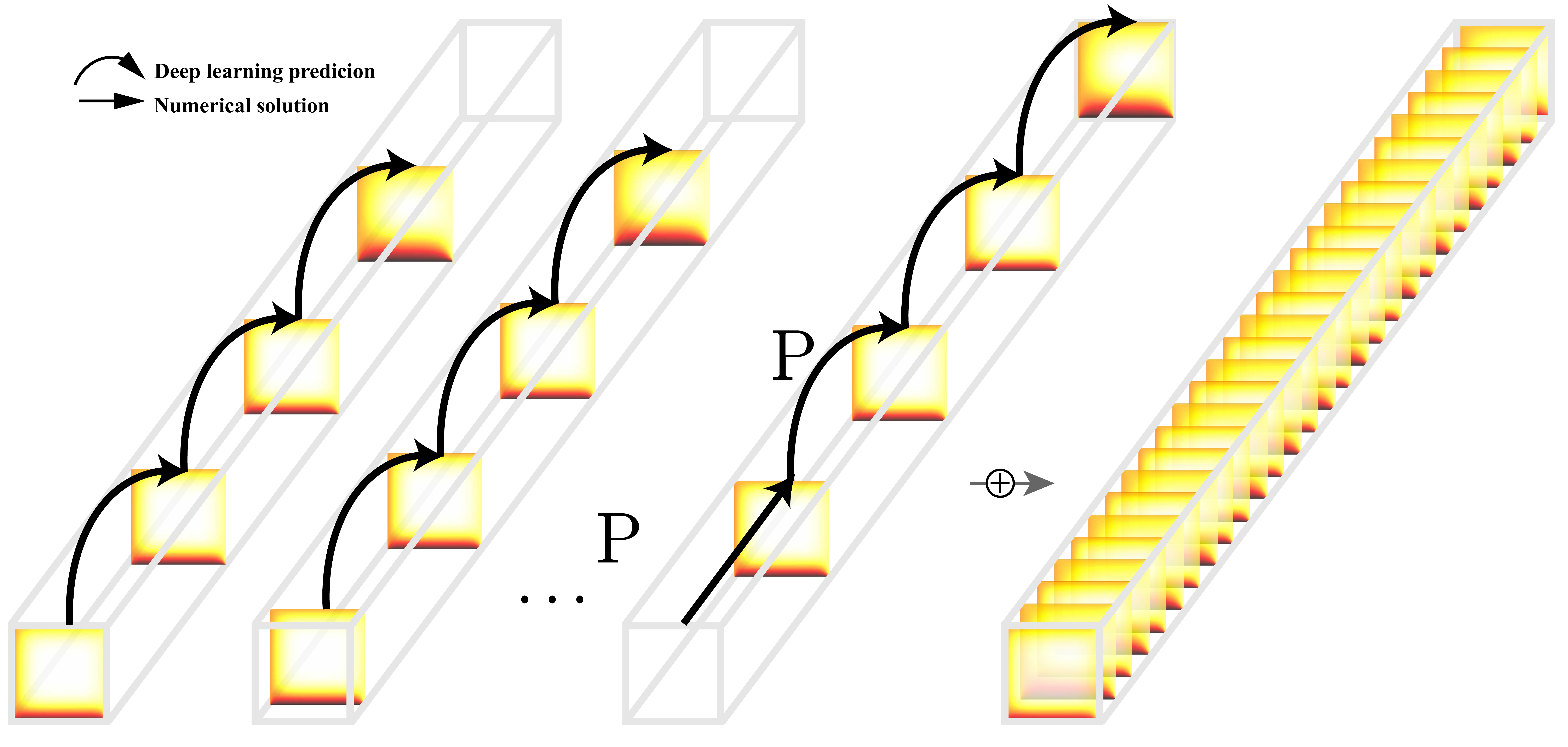}
\captionsetup{justification=centering}
\captionof{figure}{Dividing a Time-dependent Problem into chunks solved at time interval  $P$  and then recombining for the full solution }  
\end{center}
 
This section proposes a hybrid method to accelerate the PDE solution by exploiting our model ability to predict a given input's solution after multiple time steps at once. We denote  the prediction multiple steps with $P$   , for a given time-dependent  solution  $X(t =0) ,X(t=1) , … X(t=L)$ , we can divide the solution into $P$ chucks $C_0,…C_P$ as the following; 

\begin{gather*}
X(0) , X( P )  , X(2P ) , ... X(\lfloor \frac{L}{P}\rfloor P   ) \notag \\
X(1),X(P+1) ,X(2P+1) , ... X(\lfloor \frac{L-1}{P} \rfloor P + 1) \notag \\ 
\vdots \notag \\
X(P-1) , X(2P-1) , ...,  X(\lfloor \frac{L-P-1}{P} \rfloor P + P-1) 
\end{gather*}

We then train our model to predict $X(t_i+P) $from a given $X(t_i)$. Repeating this process recursively, we can see that each chunk can be solved separately and recombined later. This process is parallelizable into $P$ processes. Moreover, under the assumption that prediction error from any time instant $t_i$ to $t_i + P$ is similar, we can hypothesize that this method also reduces the recursive error propagation since each chunk requires a maximum of $\lfloor \frac{L}{P}\rfloor$ predictions instead of $L$ predictions in non chunked form. Additionally, depending on the size of the solution input and the number of the network parameters, The single prediction step inference time can be less than the numerical solution step time, thus offers extra speed. Moreover, since each chunk represents the solution at step $P$, we can investigate a computationally costly solution by solving the problem by partially processing some chunks.

\subsection{Training}

We Trained our model on generated Data by utilizing the solver implemented in NumPy-Numba.We divide our dataset into batches of equal sizes of 32 samples .
For each batch we generate random initial solution step $t_i$  such that within each batch the input and target solutions is selected at the same time $t_i $ and $t_i + P$ respectively. However each solution within the same batch has a different permutation of  six variables:Single Initial condition $IC$ , Four Boundary conditions $BC_1,BC_2,BC_3,BC_4$ , $\lambda = \frac{k \Delta t}{(\Delta x)^2}$ . 

We utilized parallel processing methods to increase the speed of the data generation process.

\begin{figure*}
\centering
   \includegraphics[width=6in]{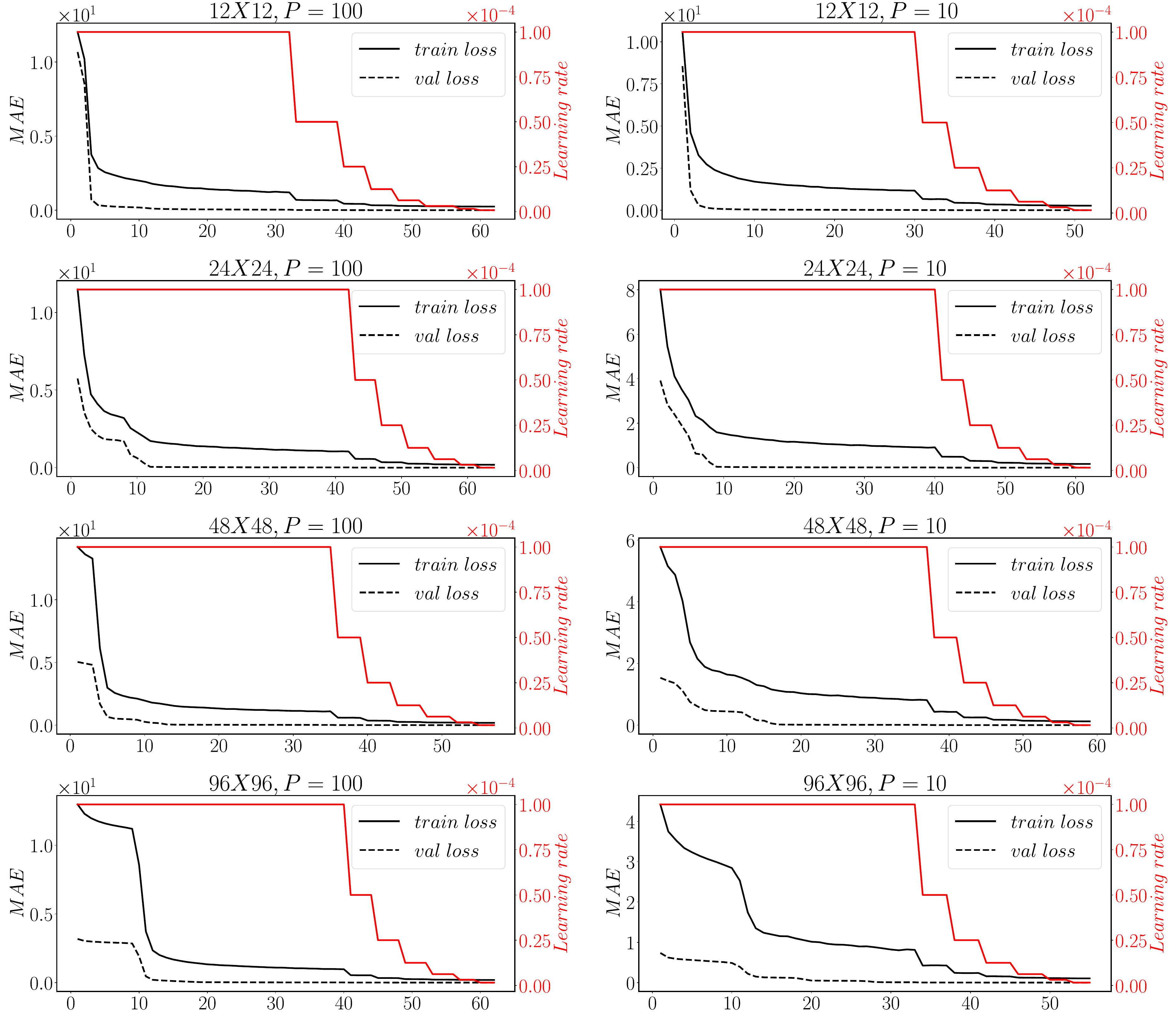}
   \captionsetup{justification=centering}
    \captionof{figure}
    {
    Loss plots for grid size $\in [12 \times 12,24 \times 24,48 \times 48,96 \times 96] \times$ prediction step $\in [10,100]$
    }   
\end{figure*}

\section{Experiment setup}
\subsection{Physics solver}
We implemented a numerical solver based on the alternating direction implicit  in python discussed in the background section. We used the Python NumPy library to implement the solver and accelerated the solution by utilizing Numba. Numba is LLVM-based Python JIT compiler \cite{lam2015numba}  we utilized to increase our solver's speed  \cite{asem000_2020_4032190}  for a single solution. Numba, in general, can offer speeds similar to other compiled languages like C++ or FORTRAN \cite{lam2015numba}. We chose the NumPy-Numba combination to implement our solver due to the convenience of integration with the rest of the codebase that became instrumental during the experimentation phase.

% \includegraphics[width=3in]{DiffusionNet/Numba-NumPy solution.pdf}
% \captionsetup{justification=centering}
% \captionof{figure}{Solver implementation speed comparison for single solution executed for 100 iterations.}   

% \end{multicols}

\begin{figure*}
\begin{center}
   \includegraphics[width=5in,keepaspectratio]{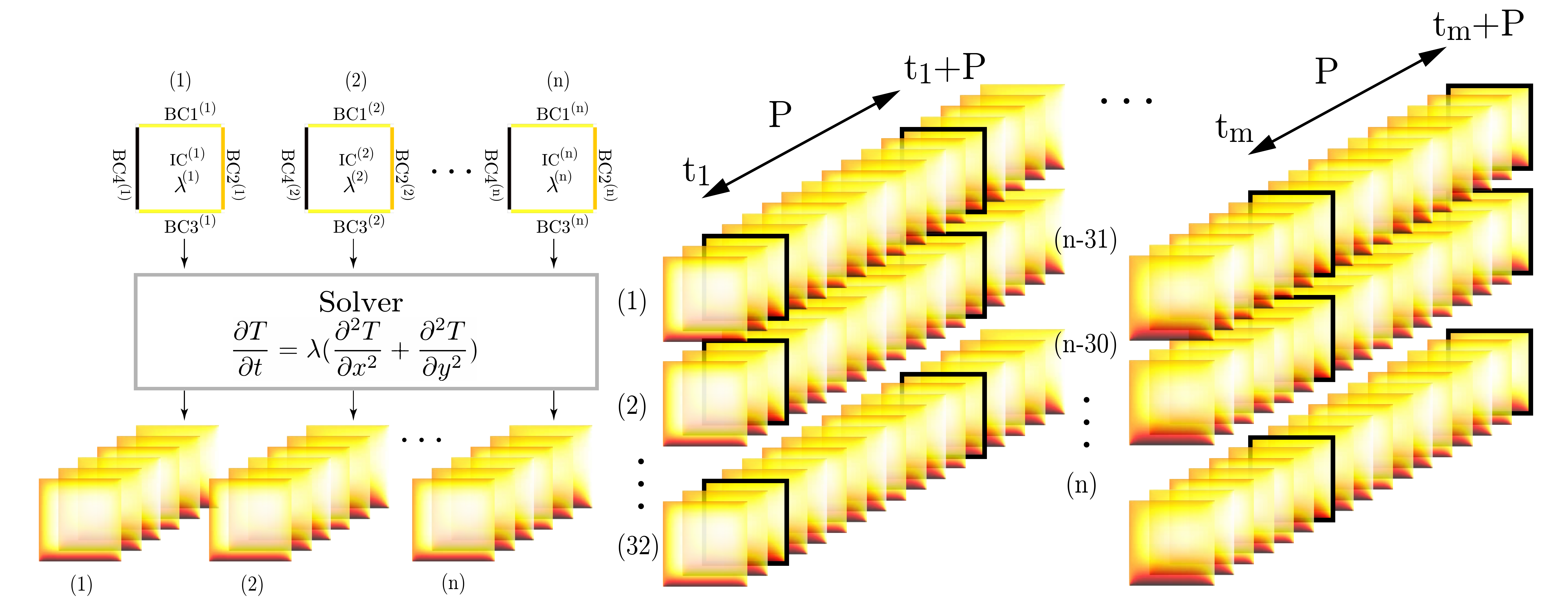}
   \captionsetup{justification=centering}
    \captionof{figure}
    {
    Training data generation for Diffusion Net.
    (Left)Data generation of N samples using random boundary conditions, initial condition, and the collective diffusion coefficient Lambda
    (Right)Random choice of input and target solution for each batch.
    }
\end{center}
\end{figure*}  

% \begin{multicols}{2}

\subsection{Data generation}
We trained the model by generating eight datasets of 10,000 batches of 32 solutions per batch. All the datasets have the same variables permutation but are solved for different prediction steps and grid sizes. The range for Boundary conditions and initial condition temperature $\in[0,100] $, and for $\lambda \in [0,1]$ . We choose a random initial timestep from a range $\in [0,1000]$ time steps. As discussed before, a random initial timestep is set for each batch, while within the batch, the solution is generated using the permutation generated from the boundary, initial condition, and $\lambda$. We generated four datasets with prediction step = 10 ( prediction of the solution after ten frames at once), and four datasets are for prediction step = 100. For each prediction step, We generated grid sizes of 12x12, 24x24, 48x48, 96x96 nodes. The MAE in loss plots is for the single prediction from $X(t_i)$ to $X(t_i+P)$ . Code for data generation and models is found at \texttt{https://github.com/ASEM000/DiffusionNet}

\begin{algorithm}[H]
\DontPrintSemicolon
\KwIn {
    $~~~(1)initial \ \& \ boundary \ conditions \ range\ (ibR)$ 
    $~~~(2)\lambda \ range \ (lR)$ 
    }
\KwOut{$Random \ Permutation \ from \ an  \ input \  range  \ $}
$ Result \leftarrow  (BC_1 , BC_2,BC_3,BC_4,IC,\lambda) $ \; 

\caption{GenerateRandomPermutation}  
\Return{Result}
\end{algorithm}

\begin{algorithm}[H]
\DontPrintSemicolon
\KwIn {\\
    $~~~(1)time \ step \ range \ (tR)$ \\
    $~~~(2)initial \ \& \ boundary \ conditions \ range\ (ibR)$ \\
    $~~~(3)\lambda \ range \ (lR)$ \\
    $~~~(4) prediction \ step (P)$
    }
\KwOut{$Data \ set$}
$ Result \leftarrow  \phi$ \; 
\For{i = 1 to batches}{
$ t_0 \leftarrow \texttt{GenerateRandomTimeStep(tR)}$\;
$ t_1 \leftarrow  t_0 + P$\;    
$ t \leftarrow [t_0,t_1]$\;
$Batch \ result \leftarrow \phi$\; 

\For{j =1 to batch size}{
$ A \leftarrow \texttt{GenerateRandomPermutation(tR)}$\;
$ Solution \leftarrow \texttt{SolveConduction(t,A)}$\;  
$Batch \ result[j] \leftarrow Solution$\; 
}
$Result[i] \leftarrow Batch \ result$\;
}

\caption{Data generation}  
\Return{Result}
\end{algorithm}

\begin{figure*}
\begin{center}
   \includegraphics[height=5in,height=8in,keepaspectratio]{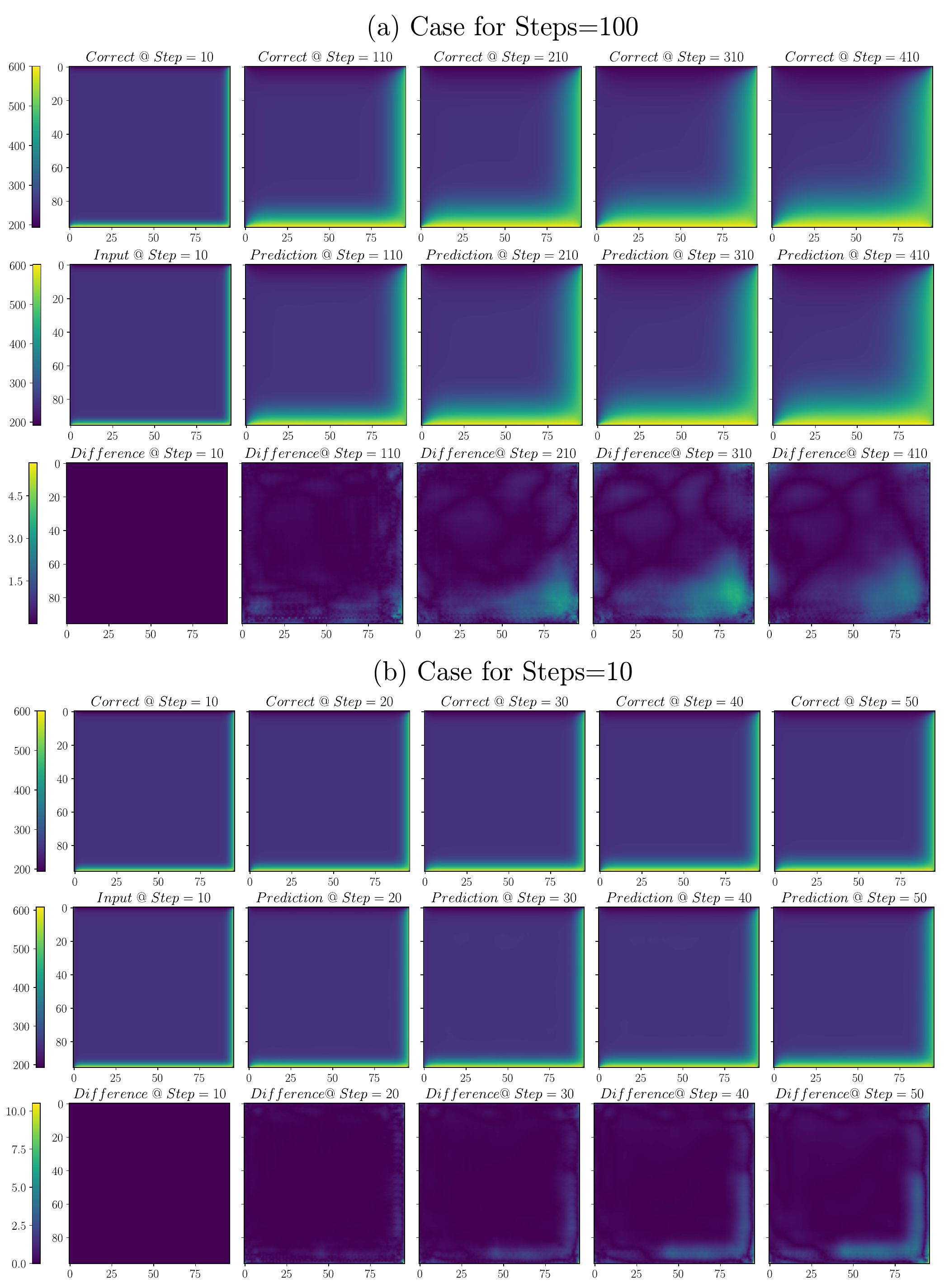}
   \captionsetup{justification=centering}
    \captionof{figure}
    {
    Comparison of Deep learning solution to Numerical solution for $96 \times 96$ grid.Deep learning method predict solution at 100 and 10 steps per prediction respectively by recursive input.
    (a) Solving for  permutation of $(BC_1,BC_2,BC_3,BC_4,IC,\lambda) = (600,500,194,248,254,0.27047)$ from starting step $=10$ for \textbf{100} steps per prediction (b) Solving for  same permutation of $BC_1,BC_2,BC_3,BC_4,IC,\lambda$ as (a) from starting step $=10$ for \textbf{10} steps per prediction
    }
\end{center}
\end{figure*}  

% \begin{multicols}{2}
\subsection{Results}

\subsubsection{Model testing}
We tested our model on 10 data sets each of 1000 batches with batch size = 1. since  each batch is set to random initial random time, we have 1000 solution over a range of six variables: $BC1, BC2, BC3, BC4, IC,\lambda,t_i$. We tested the datasets of grids sizes $\in [12 \times 12, 24 \times 24,48 \times 48, 96 \times 96]$ using models trained on the respective grid size; we then use the model trained on the $96 \times 96$ grid size to test on a dataset of size $192 \times 192$ that has never been trained on such size. All the datasets are generated using the same permutation of boundary conditions, initial condition,$\lambda$, and initial time step $t_i$; the only difference is their grid size. The mean and standard deviation of test datasets is tabulated along with the prediction's mean absolute error. We note a similar MAE for the grid size with the trained model, while the MAE of the dataset of size $192 \times 192$ is higher but still acceptable in respect to the mean and the standard deviation of the dataset.

\begin{center}
 \begin{tabular}{||c c c c c ||} 
 \hline
 ~ & ~ & Step (P)  = 10 & ~ & ~ \\
 \hline
 Grid size & MAE & Data mean & Data $\sigma$ & Trained \\ [0.5ex] 
 \hline\hline
 12 $\times$ 12  & 0.28244 & 501.25 & 172.05 & $\checkmark$ \\ 
 \hline
 24 $\times$ 24  & 0.17526 & 502.92 & 158.44 & $\checkmark$ \\ 
 \hline
48 $\times$ 48 & 0.12151 & 502.71 & 152.32 &  $\checkmark$ \\ 
 \hline
 96 $\times$ 96 & 0.10718 & 500.87 & 146.9 &  $\checkmark$ \\ 
 \hline
  192 $\times$ 192 & 0.80069 & 500.21 & 134.30 &  \\ 
 \hline
\end{tabular}
\end{center}

\begin{center}
 \begin{tabular}{||c c c c c ||} 
 \hline
 ~ & ~ & Step (P)  =100 & ~ & ~ \\
 \hline
 Grid size & MAE & Data mean & Data $\sigma$ & Trained \\ [0.5ex] 
 \hline\hline
 12 $\times$ 12  & 0.23627 & 501.72 & 172.17 & $\checkmark$ \\ 
 \hline
 24 $\times$ 24  & 0.18410 & 502.46 & 158.64 & $\checkmark$ \\ 
 \hline
48 $\times$ 48 & 0.18058 & 503.19 & 152.80 &  $\checkmark$ \\ 
 \hline
 96 $\times$ 96 & 0.19207 & 502.02 & 148.42 &  $\checkmark$ \\ 
 \hline
  192 $\times$ 192 & 6.20200 & 501.12 & 135.86 &  \\ 
 \hline
\end{tabular}
\end{center}

\subsubsection{Speed comparisons}
In this section, we examine the computation speed, grid size, and MAE relations between the JIT-compiled physics solver and DiffusionNet. We used Nvidia RTX 2080Ti and Intel® Xeon® E-2136 Processor to perform this experiment. The Physics solver utilizes the processor, while the deep learning model inference relies on the GPU. We benchmark the performacne of a \textbf{single} chunk .

First, we investigate the input size on performance. We used the model trained on a grid of size $96 \times 96$ to predict 100 steps at once ($P=100$) to test on input sizes of multiples of 96. We fixed the variables $(BC_1,BC_2,BC_3,BC_4,IC,\lambda) = (600,500,194,248,254,0.27047)$ . 
The objective of both the deep learning model and the numerical solver is to compute the solution at time step= 100. the deep learning solves the problem in a single prediction to reach step =100 while the physics solver completes the solution after 100 iterations.We repeat the same procedure for $P=10$ ,In both cases we observe speed gains as demonstrated by \ref{sizespeedup}.The solution is predicted with the model that is trained on  grid size = $96 \times 96$

\begin{center}
\includegraphics[width=\columnwidth]{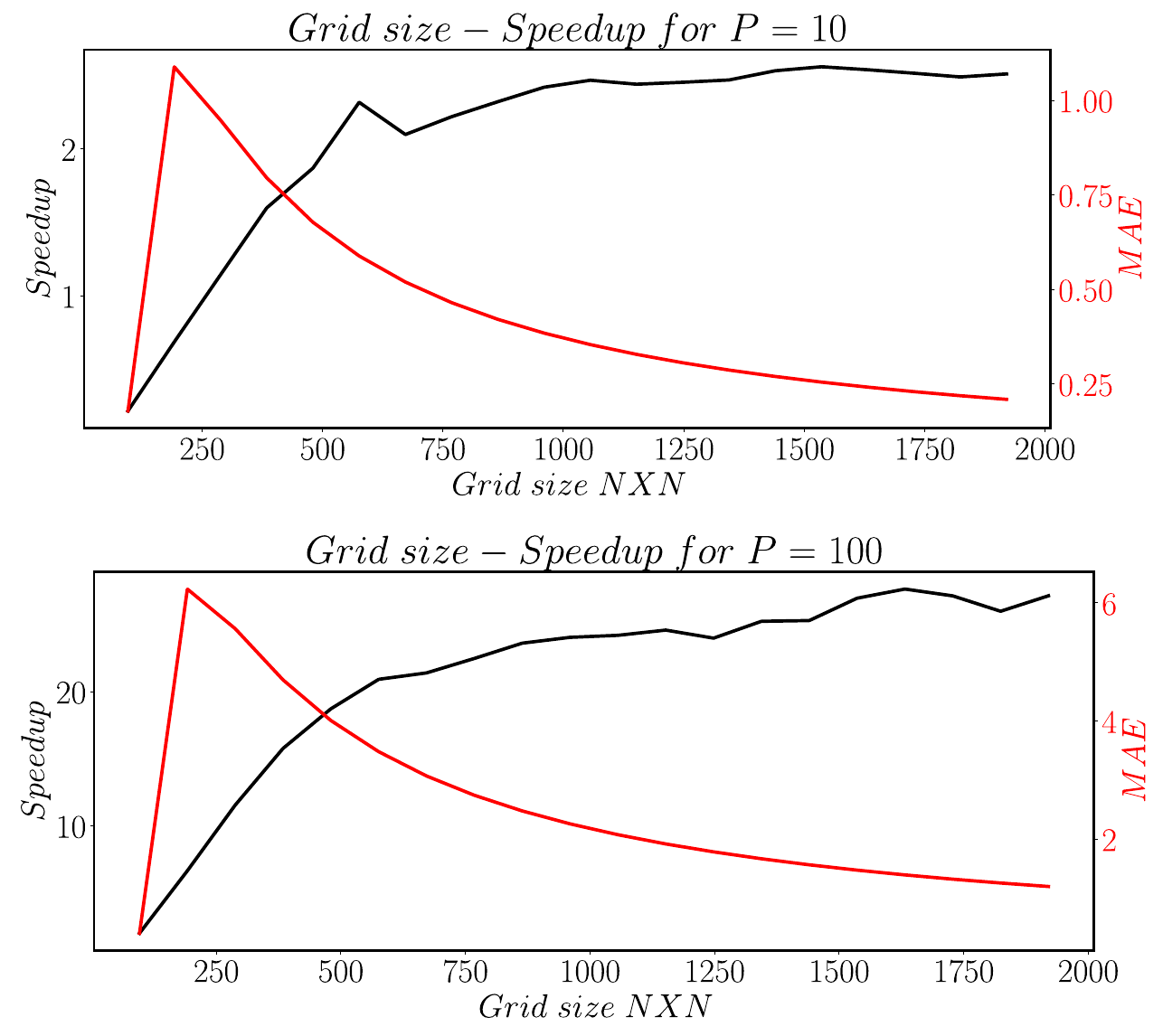}\label{sizespeedup}
\captionsetup{justification=centering}
\captionof{figure}{Comparison of input grid size to the ratio of Execution time of numerical solver to Deep learning inference  $P=10,P=100$  top and bottom respectively.}
\end{center}

Second, we investigate the effect of iterations on speedup and MAE. We fix the grid size = $960 \time 960$ and run the experiment with iterations from 100 steps up to 2000 steps. We solve the problem with the same permutation as before.We repeay the same procedue for $P=10$ , for iterations from 10 to 200.We observe $\approx$ constant speed gain relative to increasing iterations .The solution is predicted with the model that is trained on  grid size = $96 \times 96$

\begin{center}
\includegraphics[width=\columnwidth]{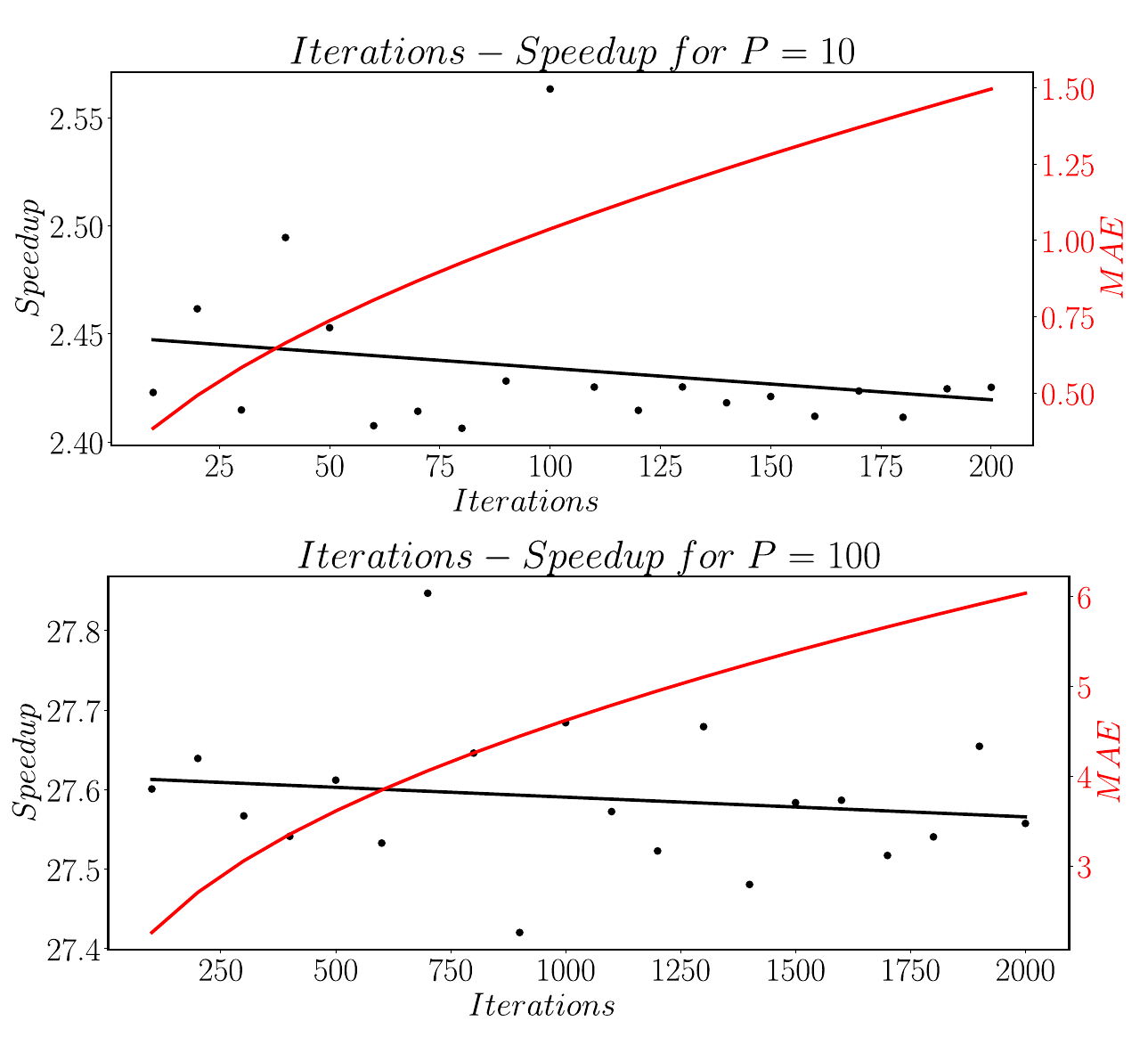}
\captionsetup{justification=centering}
\captionof{figure}{(Top) Comparison of iteration steps to the ratio of Execution time of numerical solver to Deep learning inference  for $P=10,P=100$ top and bottom respectively for grid $960 \times 960$. }
\end{center}

From both plots in this section , We can observe the significant potential speed up for the calculation of large size grids for long iterations.

\section{One dimensional time-dependent partial differential equation}

This section applies DiffusionNet to solve the inviscid Burgers' equation as an example of a one-dimensional time-dependent partial differential equation. Inviscid Burgers' equation can be considered as a prototype for equations that develop discontinuities. The advective form of the equation is

\begin{equation}
    \frac{\partial u}{ \partial t} + u \frac{\partial u}{\partial x} = 0
\end{equation}

In this section, We solve the time-dependent inviscid Burgers equation. Our objective is to investigate the reduction of error against WENO5 and FiniteNet\cite{stevens2020finitenet}.

We used the same model architecture and hyperparameters without any fine-tuning for the current problem to test the model's flexibility. We used the data provided by \cite{stevens2020finitenet} for the inviscid Burgers' equation. The data is 2000 sample solutions of Burgers' equation with random initial conditions solved for 100 steps. We trained our model on the full 2000 samples. The mean and standard deviation of the data is approximately 1.48 and 0.79, respectively. Since our model accepts the input as a 4D tensor, We reshaped the sample one spatial dimension into two spatial dimensions. We then standardized the training data by subtracting the mean and dividing by the standard deviation. We then padded the solution with zeros to have a suitable input dimension for the model. We trained the model for $P=10 $for less than 100 epochs; until the training mean absolute error stalled to $ (MAE) \approx 1.5e-3 $. 

\begin{center}
\includegraphics[width=3in]{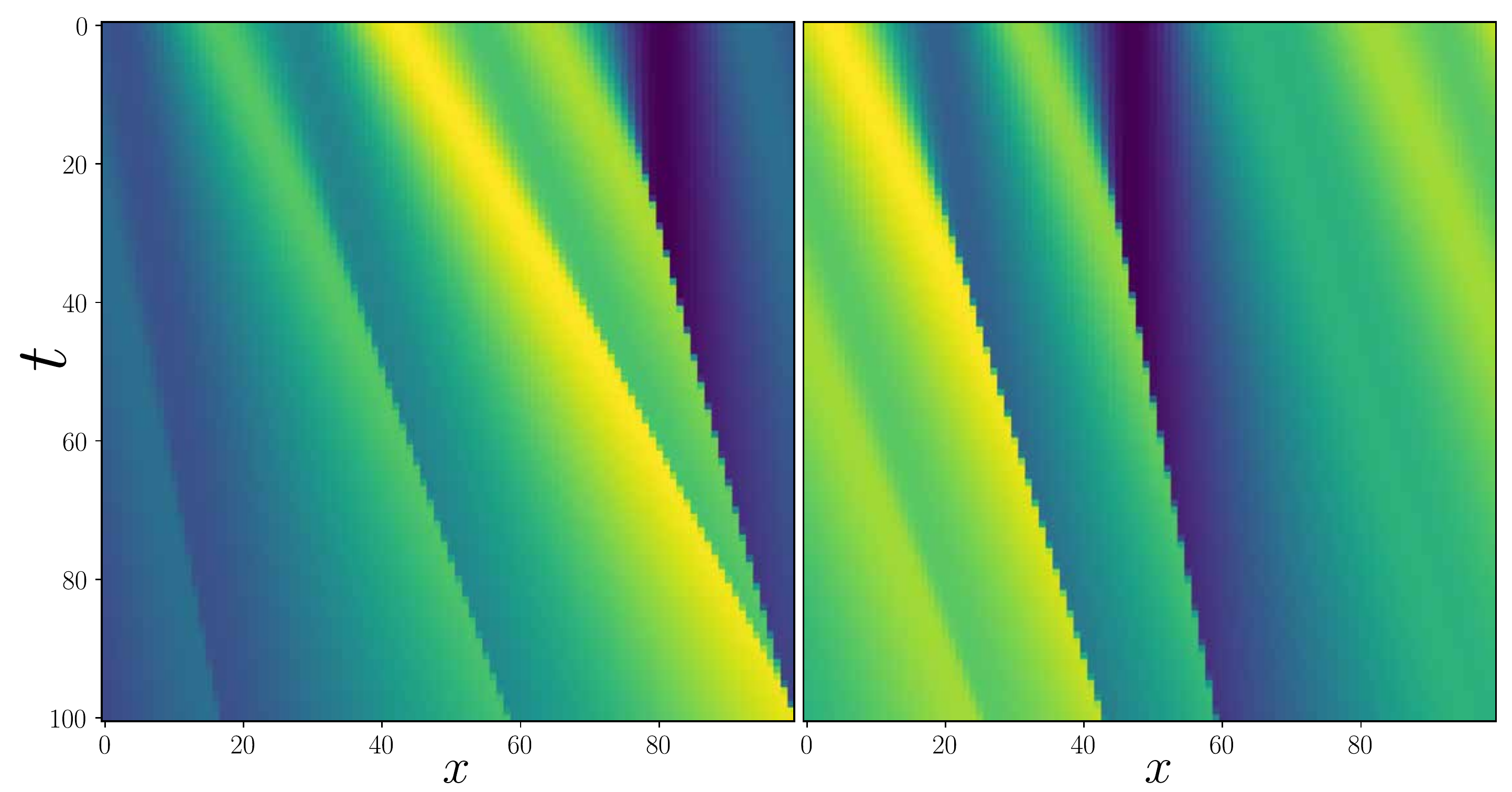}
\captionsetup{justification=centering}
\captionof{figure}{Sample solutions of inviscid Burgers' equation}     
\end{center}

Similar to \cite{stevens2020finitenet}, we tested our model on 1000 simulations provided by \cite{stevens2020finitenet} test script and then compared our model error to WENO5 and FiniteNet errors. Since our model predicts multiple steps at once ( in our case we set $Pt = 10$ ), we divided the solution into ten chunks  $C1,C2,...C10$ = $(X(0),X(10),X(20)...X(100))$ , $(X(1),X(11),X(21)...X(91))$. , .. $(X(9),X(10),X(20)...X(99))$ , We then computed the error of each chunk $C1,C2,...C10$ and the total error of the solution $\Sigma C_i$ Mean squared error to the exact solution.Each chunk initial starting point $t>0$ $X(1) ,...X(9)$ is calculated from WENO5. We investigated the MSE error for each chunk and full solution among DiffusionNet, FiniteNet, and WENO5. We investigated the error of a single chunk compared to combined chunks solution(i.e., Full solution ); In the sample frequency plots we observe that the number of samples with lower error than WENO5 is more than any of the single chunks. We can conclude that our division strategy mitigated the recursive prediction error by simply reducing the recursive predictions required. The full solution obtained by DiffusionNet achieves lower error on 947 out of 1000 samples.

% \end{multicols}

\begin{figure*}
\includegraphics[width=\textwidth]{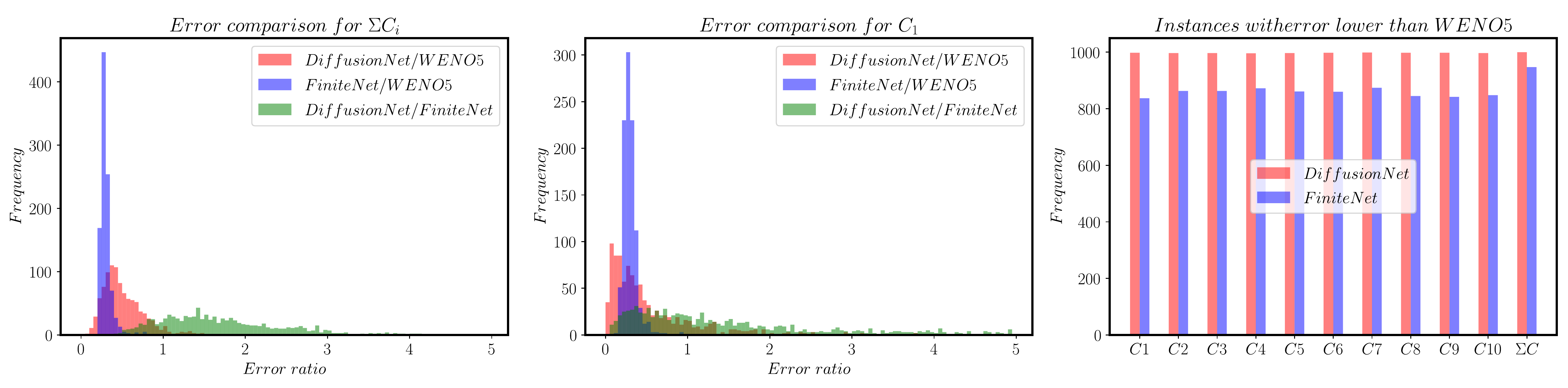}
\captionsetup{justification=centering}
\label{fig:hist}
\captionof{figure}
{
\emph{Left}Error ratios with respect to exact solution.
Ratio of DiffusionNet MSE to WENO5 MSE , FiniteNet MSE to WENO5 MSE and DiffusionNet MSE to FiniteNet MSE respectively for the full solution steps $X(0),X(1),...X(100)$. 
\emph{Center} Ratio of DiffusionNet MSE to WENO5 MSE , FiniteNet MSE to WENO5 MSE and DiffusionNet MSE to FiniteNet MSE respectively for a solution chunk at 10 steps $X(0),X(10),X(20)...X(100)$. 
\emph{Right} Sample Frequency of FiniteNet and DiffusionNet with lower error than WENO5 for solution at 10 steps with different starting point,where $C1,C2,...C10$ denote the solution chunks   $(X(0),X(10),X(20)...X(100))$ , $(X(1),X(11),X(21)...X(91))$. , .. $(X(9),X(19),X(29)...X(99))$ respectively and  $\Sigma C$ denotes the full solution steps
$(X(0),X(1),X(2)...X(100))$.
}   

\end{figure*}

\begin{figure*}
   \includegraphics[height=8in]{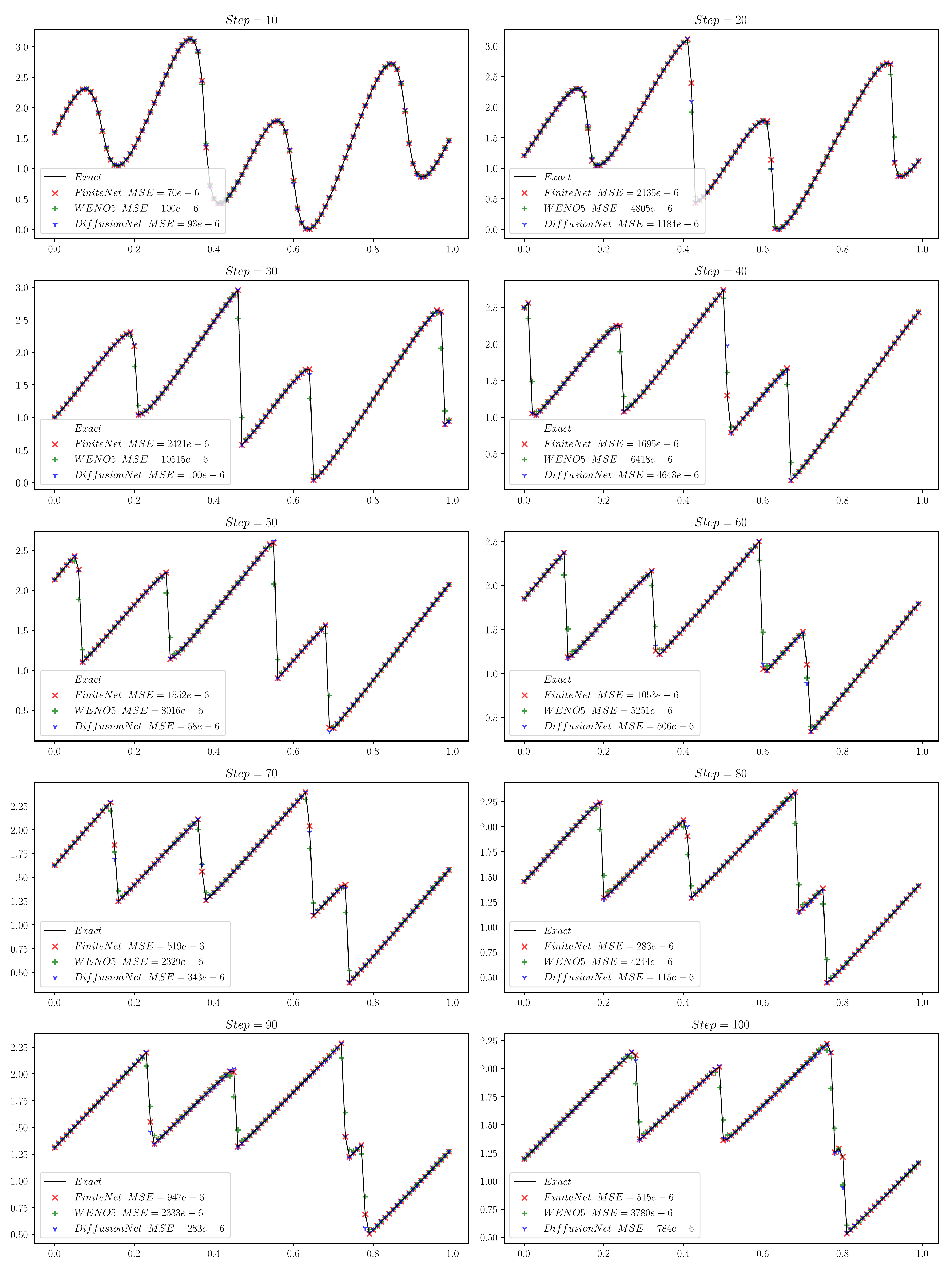}\label{diffusionfiniteplot}
   \centering
   \captionsetup{justification=centering}
    \captionof{figure}
    {
    Comparison of FiniteNet, WENO5 and DiffusionNet MSE to the exact solution of inviscid burgers' equation for a sample solution at different time steps
    }   
\end{figure*}

% \begin{multicols}{2}

\section{Steady state heat conduction}
In this section, we apply DiffusionNet to solve for the Laplace equation; we chose the two-dimensional steady-state heat conduction as our problem of choice to compare with the existing literature. for our solution scheme, a steady-state solution is framed as a single prediction step problem. 

The temperature distribution is governed by the Laplace equation (\ref{steady1}).Laplace equation can be solved by discretizing the solution for evenly spaced grid $\Delta  x = \Delta y$ using (\ref{steady2})

\begin{gather}
\frac{\partial^2 T }{\partial x^2} + \frac{\partial^2 T }{\partial y^2}  = 0 \label{steady1} \\ 
T_{i,j} = \frac{T_{i+1,j} + T_{i-1,j} + T_{i,j+1}+T_{i,j-1}}{4}  \label{steady2}
\end{gather}

We use data provided by \cite{edalatifar2020using} , which is composed of different geometries solutions of the Laplace equation. The data is composed of $64 \times 64$ images, the input image has two channels, and the output has one channel. We preprocessed the data by standardization. The mean and the standard deviation is calculated from the training dataset. The mean and standard deviation of the training data is 0.17192 and 0.36882, respectively.

We trained the model for less than 100 epochs until the loss stalled. We used our model without any modification in hyperparameters or architecture.\cite{edalatifar2020using} introduced different loss functions and then trained their network for each of the loss functions introduced. We compare our model error to \cite{edalatifar2020using} mean absolute error on the same train, validation, and test data.
We observe that DiffusionNet outperforms the model with Net MMaSE for train, validation, and test with nearly half the error. For NetMSE, DiffusionNet outperforms the model the validation and test cases. We then conclude our model's ability to solve the Laplace equation with different geometry with no changes in the model or hyperparameters at high fidelity.

\begin{center}
 \begin{tabular}{||c l l l||} 
 \hline
 ~ & Train & Validation & Test \\ [0.5ex] 
 \hline\hline
 DiffusionNet & 0.000715 & \textbf{0.000728} & \textbf{0.000727} \\ 
 \hline
 Net MSE \cite{edalatifar2020using} & \textbf{0.00061} & 0.000796 & 0.000791  \\
 \hline
 Net MMaSE\cite{edalatifar2020using} & 0.0014 & 0.0015 & 0.0015 \\
 \hline

\end{tabular}
\end{center}

\begin{center}
\includegraphics[width=\columnwidth]{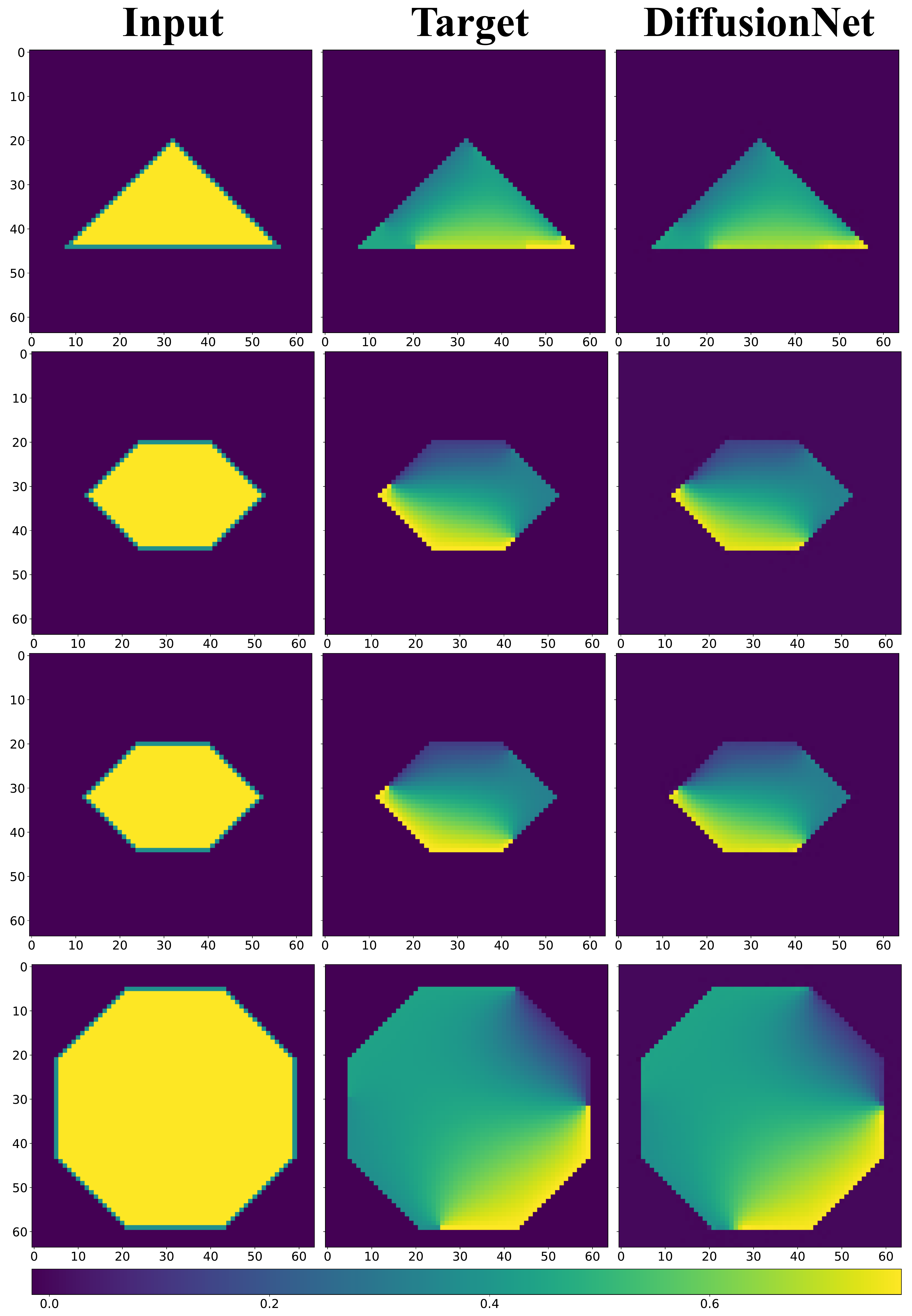}\label{laplace}
\captionsetup{justification=centering,margin=1cm}
\captionof{figure}{The solution of a test sample of input and target solution for steady state heat equation.\emph{Left} Input image with the initial conditions.\emph{Center} Target output.\emph{Right} DiffusionNet output.}    
\end{center}

\section{Conclusion}

In this paper, we presented our deep learning framework to solve and accelerate the time-dependent partial differential equation's solution by dividing the solution timesteps into chunks that can be solved independently. We solved the transient heat conduction in two spatial dimensional, time-dependent inviscid Burgers' equation and the steady heat conduction. We used the same architecture and training scheme to solve the above problems.
We compared the speedup over the full solution by benchmarking the execution time of solving a single chunk against the JIT-Compiled physics solver solution. We investigated the relationship between speed gains, MAE, and iterations and found potential significant speed gains at large grid sizes for long iterations. We then solved the inviscid burgers equation To examine the framework prediction error propagation.
, we observe that our framework can mitigate recursive prediction error; by comparing it to the baseline algorithm to solve the equation, our model achieved lower error on 94.7\% of the cases.
Moreover, To demonstrate our framework's applicability for time-independent problems, we applied our model to solve the steady-state heat equation. We reduced the MAE error on the problem of steady-state conduction compared to the literature. The three experiments prove that Faster and error reducing deep learning-based solvers are possible.

\newpage

\bibliographystyle{plain}  


\bibliography{bib.bib}

\begin{thebibliography}{10}

\bibitem{asem000_2020_4032190}
ASEM000.
\newblock {ASEM000/High-performance-ADI-solver-using-numba 1.0}, September
  2020.

\bibitem{carnahanj}
B~Carnahan and HA~Luther.
\newblock J. 0. wilkes. 1969. applied numerical methods.

\bibitem{chapra2010numerical}
Steven~C Chapra, Raymond~P Canale, et~al.
\newblock {\em Numerical methods for engineers}.
\newblock Boston: McGraw-Hill Higher Education,, 2010.

\bibitem{edalatifar2020using}
Mohammad Edalatifar, Mohammad~Bagher Tavakoli, Mohammad Ghalambaz, and Farbod
  Setoudeh.
\newblock Using deep learning to learn physics of conduction heat transfer.
\newblock {\em Journal of Thermal Analysis and Calorimetry}, pages 1--18, 2020.

\bibitem{farimani2017deep}
Amir~Barati Farimani, Joseph Gomes, and Vijay~S Pande.
\newblock Deep learning the physics of transport phenomena.
\newblock {\em arXiv preprint arXiv:1709.02432}, 2017.

\bibitem{hochreiter1996lstm}
Sepp Hochreiter and J{\"u}rgen Schmidhuber.
\newblock Lstm can solve hard long time lag problems.
\newblock {\em Advances in neural information processing systems}, 9:473--479,
  1996.

\bibitem{hochreiter1997long}
Sepp Hochreiter and J{\"u}rgen Schmidhuber.
\newblock Long short-term memory.
\newblock {\em Neural computation}, 9(8):1735--1780, 1997.

\bibitem{DBLP:journals/corr/HuangLW16a}
Gao Huang, Zhuang Liu, and Kilian~Q. Weinberger.
\newblock Densely connected convolutional networks.
\newblock {\em CoRR}, abs/1608.06993, 2016.

\bibitem{lam2015numba}
Siu~Kwan Lam, Antoine Pitrou, and Stanley Seibert.
\newblock Numba: A llvm-based python jit compiler.
\newblock In {\em Proceedings of the Second Workshop on the LLVM Compiler
  Infrastructure in HPC}, pages 1--6, 2015.

\bibitem{lienhard2005heat}
IV~Lienhard and H~John.
\newblock {\em A heat transfer textbook}.
\newblock phlogiston press, 2005.

\bibitem{ranade2020discretizationnet}
Rishikesh Ranade, Chris Hill, and Jay Pathak.
\newblock Discretizationnet: A machine-learning based solver for navier-stokes
  equations using finite volume discretization.
\newblock {\em arXiv preprint arXiv:2005.08357}, 2020.

\bibitem{sharma2018weakly}
Rishi Sharma, Amir~Barati Farimani, Joe Gomes, Peter Eastman, and Vijay Pande.
\newblock Weakly-supervised deep learning of heat transport via physics
  informed loss.
\newblock {\em arXiv preprint arXiv:1807.11374}, 2018.

\bibitem{stevens2020finitenet}
Ben Stevens and Tim Colonius.
\newblock Finitenet: A fully convolutional lstm network architecture for
  time-dependent partial differential equations.
\newblock {\em arXiv preprint arXiv:2002.03014}, 2020.

\bibitem{xingjian2015convolutional}
SHI Xingjian, Zhourong Chen, Hao Wang, Dit-Yan Yeung, Wai-Kin Wong, and
  Wang-chun Woo.
\newblock Convolutional lstm network: A machine learning approach for
  precipitation nowcasting.
\newblock In {\em Advances in neural information processing systems}, pages
  802--810, 2015.

\bibitem{zakeri2019deep}
Behzad Zakeri, Amin~Karimi Monsefi, and Babak Darafarin.
\newblock Deep learning prediction of heat propagation on 2-d domain via
  numerical solution.
\newblock In {\em The 7th International Conference on Contemporary Issues in
  Data Science}, pages 161--174. Springer, 2019.

\end{thebibliography}

\end{multicols}
\end{document}